\documentclass{article}

\usepackage{arxiv}

\usepackage[utf8]{inputenc} 
\usepackage[T1]{fontenc}    
\usepackage{hyperref}       
\usepackage{url}            
\usepackage{booktabs}       
\usepackage{amsfonts}       
\usepackage{nicefrac}       
\usepackage{microtype}      
\usepackage{lipsum}
\usepackage{graphicx}
\usepackage{amsmath} 
\graphicspath{ {./images/} }

\title{Automating Artifact Detection in Video Games}

\author{
Parmida Davarmanesh \\
University of Michigan, Ann Arbor\\
\texttt{pdavar@umich.edu} \\

\And

Kuanhao Jiang\\
University of Pennsylvania\\
\texttt{kuanhaoj@seas.upenn.edu}\\

\And

Tingting Ou\\
Johns Hopkins University\\
\texttt{tou2@jhu.edu}\\

\And
Artem Vysogorets\\
New York University\\
\texttt{amv458@nyu.edu}\\

\And
Stanislav Ivashkevich\\
AMD, Inc. \\
\texttt{Stanislav.Ivashkevich@amd.com}\\

\And
Max Kiehn\\
AMD, Inc.\\
\texttt{max.kiehn@amd.com}\\

\And
Shantanu H. Joshi\\
UCLA\\
\texttt{s.joshi@g.ucla.edu}\\

\And
Nicholas Malaya\\
AMD Research\\
\texttt{nicholas.malaya@amd.com}
}

\begin{document}
\maketitle

\begin{abstract}
 In spite of  advances in gaming hardware and software, gameplay is often tainted with graphics errors, glitches, and screen artifacts. This proof of concept study presents a machine learning approach for automated detection of graphics corruptions in video games. Based on a sample
of representative screen corruption examples, the model was able to identify 10 of the most commonly occurring screen artifacts with reasonable accuracy. 
 
 Feature representation of the data included discrete Fourier transforms, histograms of oriented gradients, and graph Laplacians. Various combinations of these features were used to train machine learning models that identify individual classes of graphics corruptions and that later were assembled into a single mixed experts ``ensemble'' classifier. The ensemble classifier was tested on heldout test sets, and produced an accuracy of 84\% on the games it had seen before, and 69\% on games it had never seen before.

\end{abstract}


 \maketitle

\section{Introduction}

\begin{figure}[ht]
\centering
\includegraphics[width = \linewidth]{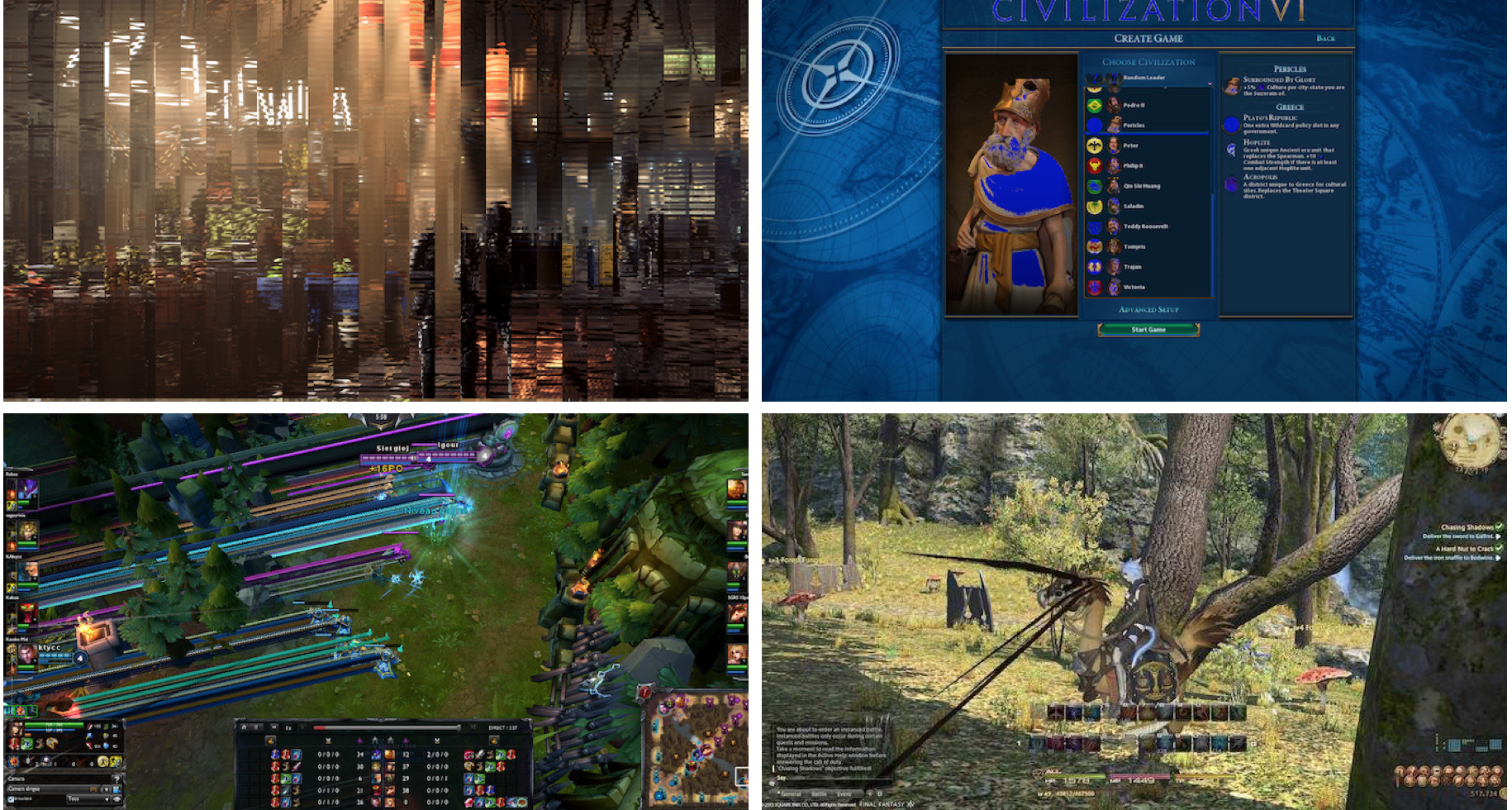}
\caption[Examples of video game artifacts]{Examples of video game artifacts (images are courtesy of AMD).}
\label{fig:bugs}
\end{figure}

More than 65\% of Americans play video games on at least one type of device \cite{stats}. Furthermore, the combined gaming industry produced a revenue of \$120B in 2019, a 4\% increase from 2018 \cite{takahashi_2020}. For these games to be immersive, gameplay must remain high-quality and error-free. While games may contain various defects and issues, graphical corruption and visual artifacts are some of the largest complaints from users \cite{taxonomy}. These artifacts typically occur due to software or hardware errors that alter visual appearance of the game or its individual frames. Figure \ref{fig:bugs} shows some examples of such corruptions observed in real games.

\noindent

 In this paper, we conducted a proof of concept study to automate the detection of graphical artifacts. This is a novel problem that, if solved, would lead to significant quality improvement and enhanced experience for consumers of images and video. Currently, artifact detection is a labor intensive process where glitches are reported individually by users who experience them \cite{glitch_arxiv}. This process is manual and time consuming, and a large number of users typically choose not to report these glitches, resulting in a number of unreported and unresolved issues. Furthermore, even if these glitches are reported in large scales, it still takes a lot of human effort to sort and catalogue these them. The first implication of automating this process is that there is no need for human intervention. Once a glitch occurs, it will be automatically captured and sent to the corresponding company for correction. Second, with increased knowledge of the source and cause of glitches, this automated software will be able to catch and correct the glitch before it is displayed to the user, resulting in an uninterrupted and smooth gameplay experience.
 To the best of the authors' knowledge, no existing work has systematically synthesized glitches in images, cataloged and labeled them, and proposed an automated solution for detecting them. \\

 The contributions of this paper are as follows: 

\begin{enumerate}
    \item Creation of the open source software, called \emph{Glitchify}, for reproducing a basket of common gaming artifacts.
    \item Generation of a labeled large dataset consisting of 50,000 normal and glitched gaming images.
    \item Using dimensionality reduction and feature extraction techniques on gaming images and building an ensemble model to classify gaming images to automate the process of artifact detection.

\end{enumerate}

\section{Gaming Artifact Creation}\label{sec:datagen}

Automating artifact detection, just like any other classification task, requires large amounts of data (e.g. labelled corrupted frames). There are no publicly available large-scale databases which provide this. Therefore, we created a large dataset of real images from game plays and injected different types of graphics artifacts  to obtain  glitched images. Specifically, we focused on the artifacts that are caused by software defects, as hardware artifacts which are highly content related are produced during rendering in the GPU and are thus much more challenging to reproduce. This data generation was implemented in an open source software \emph{Glitchify} which is publicly available at \url{https://github.com/AMD-RIPS/ST-2019}.

The prototype for our synthetic data was a limited collection of corrupted images provided by AMD that were representative of graphic artifacts observed frequently during gameplay. The \emph{Glitchify} software then artificially generated images with different types of corruption, designed to closely mimic what was observed in the sample set. Unlike artifacts shown in Figure \ref{fig:bugs}, some of the images in the sample appeared to also have content-related artifacts. These content-related glitches depend upon the objects and their interactions in scenes and are thus difficult to represent and resolve. Since these glitches would further require video segmentation and object recognition techniques for detection, they are not addressed in the paper.

\subsection{Types of Reproduced Artifacts}

To reproduce different artifacts, we defined 10 different classes of corruptions based on their appearances. Gameplay data was obtained from publicly available YouTube videos. The  dimension of the images or frames extracted from gameplay videos was set to $1920 \times 1080$ pixels. These different kinds of artifacts are described below.

\subsubsection{Shader Artifacts}
\begin{figure}[ht]
\centering
\includegraphics[scale=0.1206]{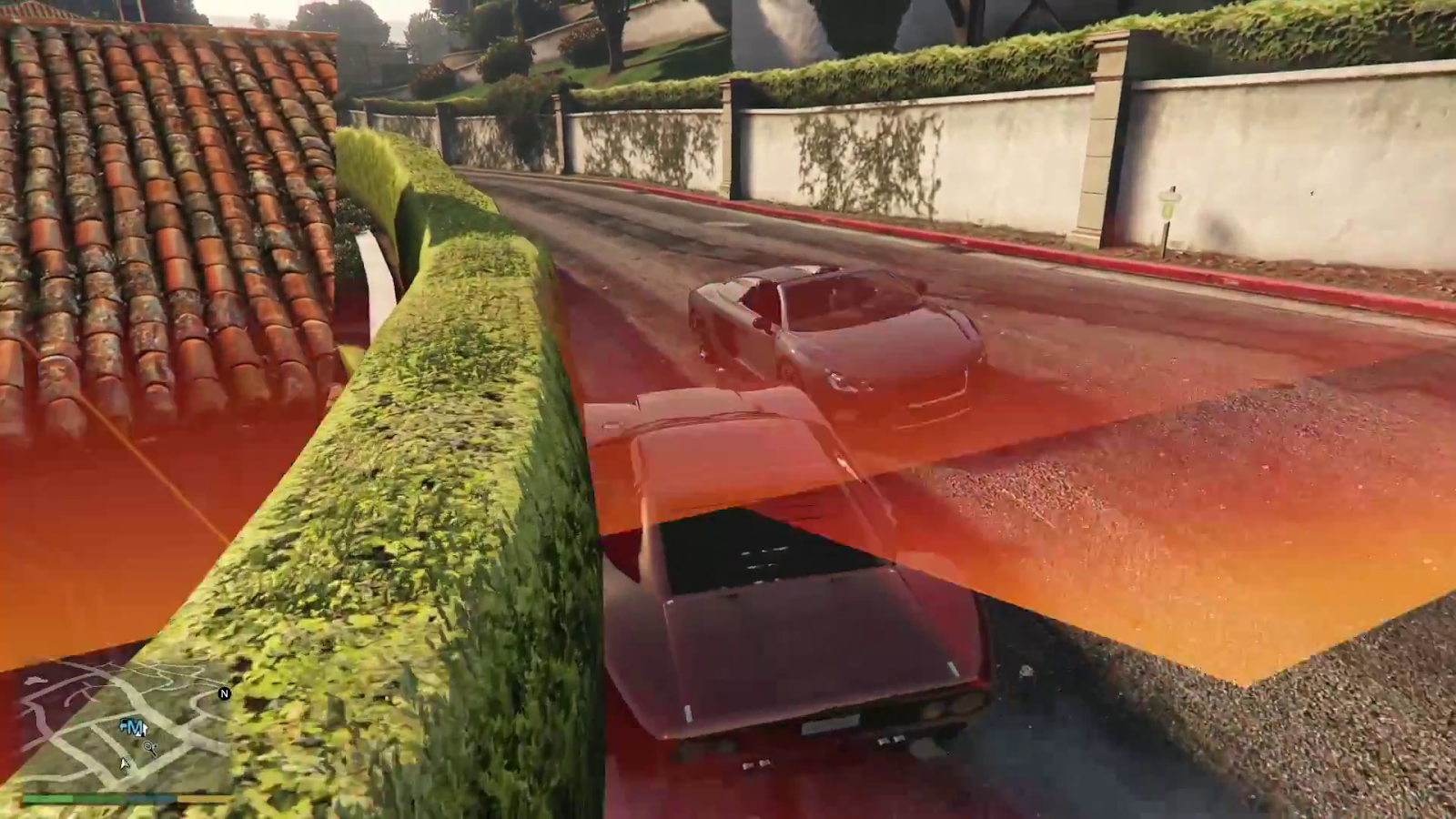}
\includegraphics[scale=.382]{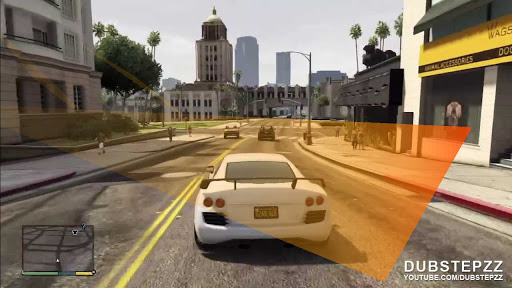}
\caption{\textit{Shader} corruption: actual (left) and reproduced with \emph{Glitchify} (right).}
\label{Shader}
\end{figure}

The shader program within a GPU performs frame rendering and determines various surface properties such as texture, reflection and lightning\cite{Shader}. For our work, \textit{shader} artifacts are marked by the presence of polygonal shapes of  different colors that either blend together or display gradual fading in certain directions. This glitch (Figure \ref{Shader}) was  reproduced by choosing a random number of points in random positions of the image as the starting point and then setting a random number of edges to form a polygon. We then chose a random color close to that of the initial point, and reassign the pixel values in that polygon by slowly changing the color and reducing the intensity of the color.

\subsubsection{Shapes Artifacts}
\begin{figure}[ht]
\centering
\includegraphics[scale=0.3636]{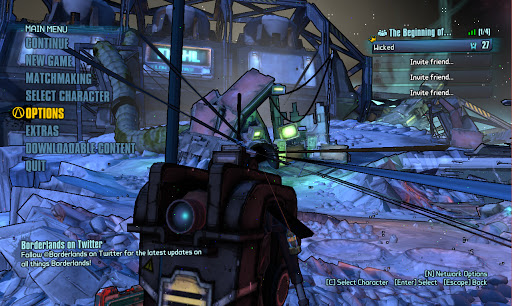}
\includegraphics[scale=0.387]{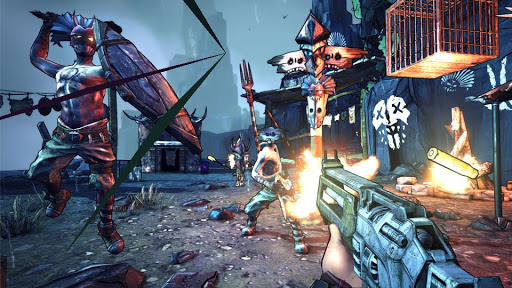}
\caption{\textit{Shapes}: actual (left) and reproduced with \emph{Glitchify} (right).}
\label{fig:shapes}
\end{figure}

Random polygonal monocolor shapes are also common in video games, especially in first-person shooting games. \textit{Shapes} artifacts tend to appear in the darker part of the frames. We reproduced this artifact (Figure \ref{fig:shapes}) by choosing a random staring point in the darkest rectangular region (of random size within a certain range) of the image, and drawing a random number of dark thin polygons out of that point.

\subsubsection{Discoloration Artifacts}
\begin{figure}[ht]
\centering
\includegraphics[scale=0.1188]{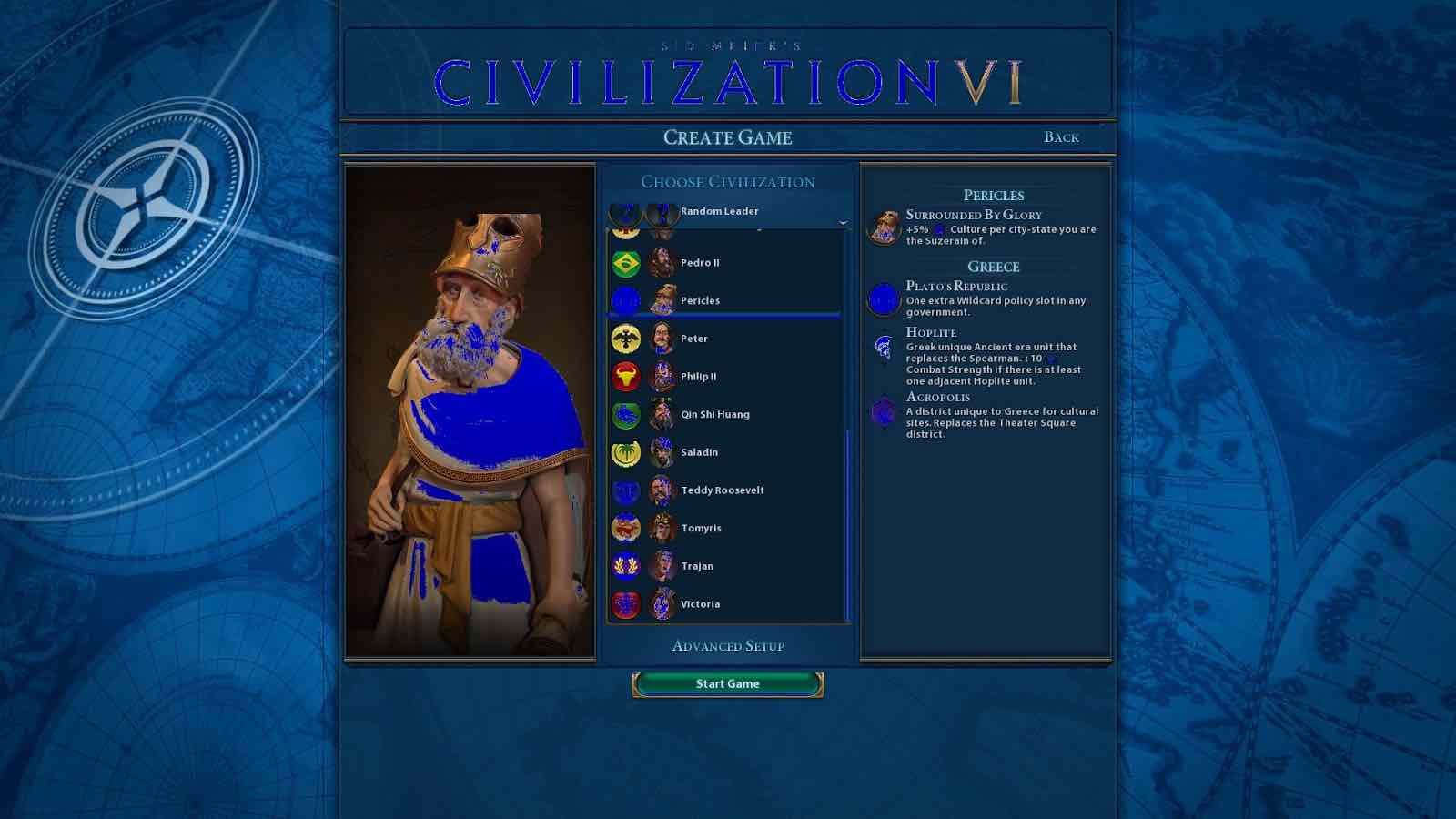}
\includegraphics[scale=0.1368]{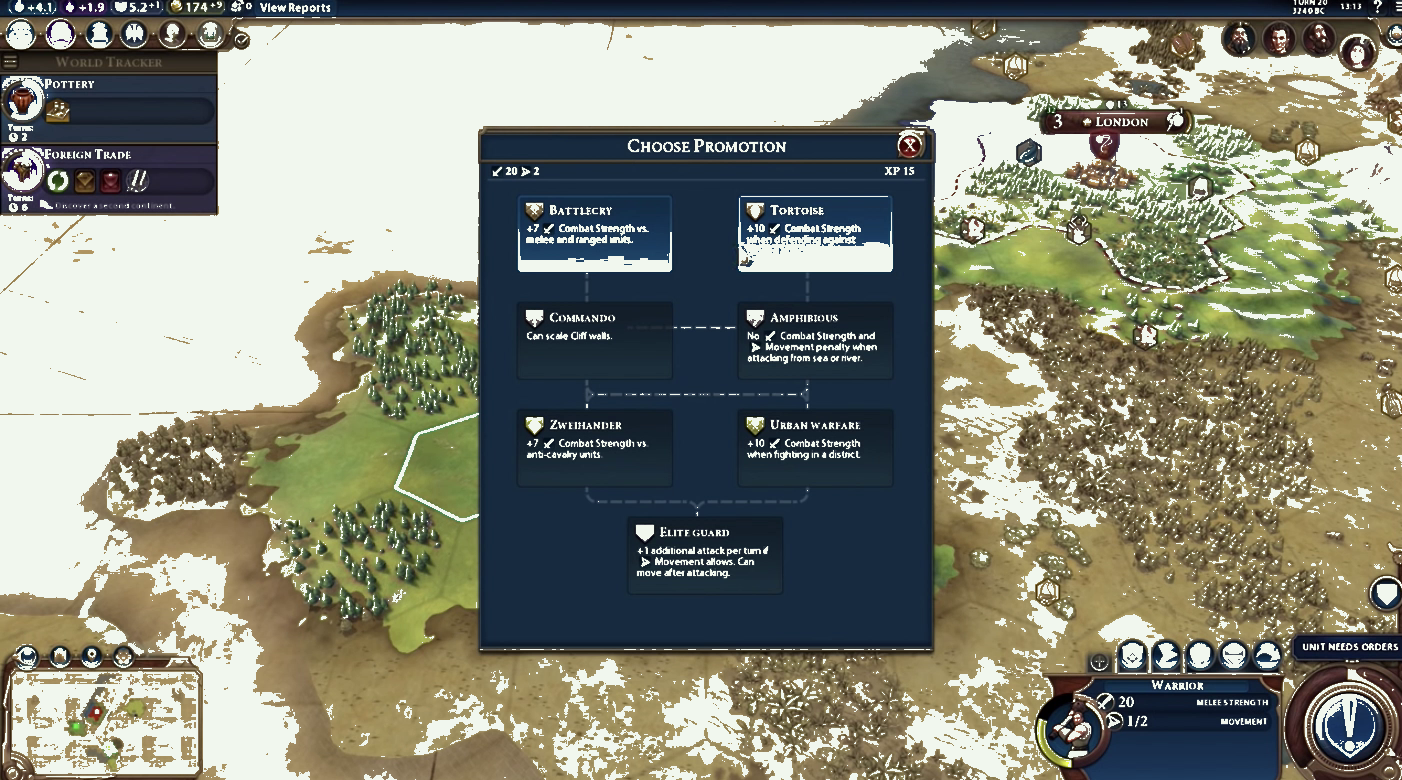}
\caption{\textit{Discoloration}: actual (left) and reproduced with \emph{Glitchify} (right).}
\label{fig:discoloration}
\end{figure}

\textit{Discoloration} artifacts manifest as bright spots in  images that are colored differently. This artifact (Figure \ref{fig:discoloration}) was reproduced by changing the color of pixels. For example, one could set the red component of the pixel to a predefined value above or below a certain threshold.

\subsubsection{Morse Code Pattern}
\begin{figure}[!h]
\centering
\includegraphics[scale=0.72]{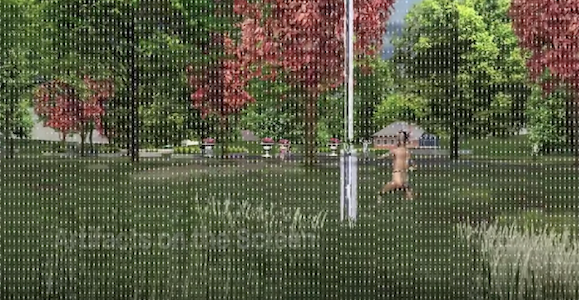}
\includegraphics[scale=0.27]{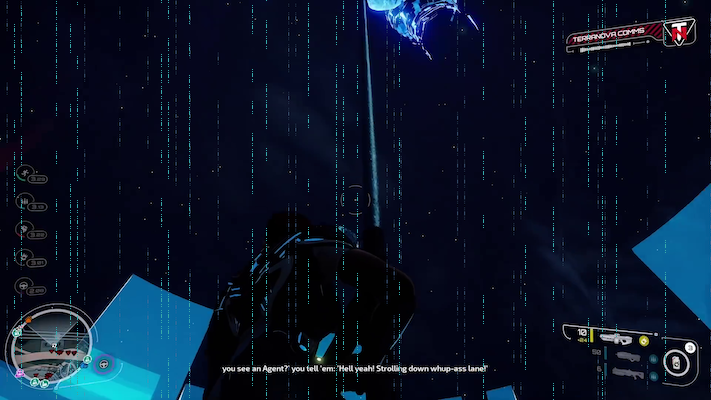}
\caption{\textit{Morse Code Pattern}: actual (left) and reproduced with \emph{Glitchify} (right).}
\label{fig:morse}
\end{figure}

The morse code pattern (shown in Figure \ref{fig:morse}) appears when memory cells on a graphic card become stuck and display their stuck values on the screen rather than displaying the true image. Running a GPU at a higher speed than it was designed for, or at a high temperature, may result in such corruption. To reproduce this type of artifact, we add morse-code-like patterns to random locations in the frame, as shown in Figure \ref{fig:morse}.

\subsubsection{Dotted Lines Artifacts}
\begin{figure}[ht]
\centering
\includegraphics[scale=.315]{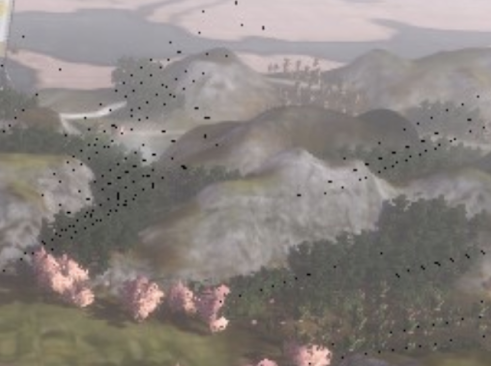}
\includegraphics[scale=.225]{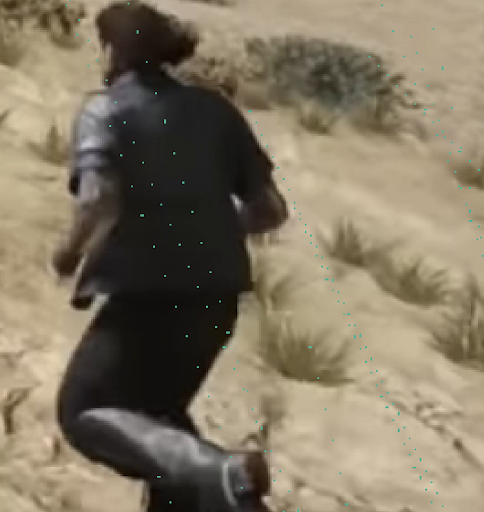}
\includegraphics[scale=.405]{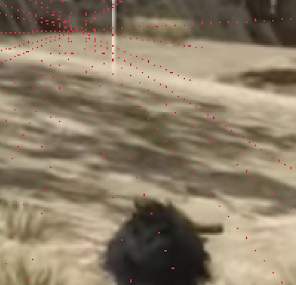}
\caption{\textit{Dotted Lines}: actual corrupted image provided by AMD (leftmost), our reproduced corrupted image with random lines of green color (middle) and radial lines of red color (rightmost).}
\label{fig:dl}
\end{figure}

\textit{Dotted Lines} artifacts are often hard to recognize unless one magnifies the corrupted images. The \textit{Dotted Lines} either have random slopes and positions in the input frames, or they can be radial lines emanating from a single point. To generate the random \textit{Dotted Lines} artifacts (Figure \ref{fig:dl}), a random color is first determined. Then $N$ \textit{dotted lines} of this color replace the original pixel values of the image, where $N$ 
is randomly chosen from a uniform distribution. We chose the starting point of each line segment such that 
it does not lie close to the edges of the image. The radial \textit{Dotted Lines} are different from the random \textit{Dotted Lines} in that they originate from a single point and $M$ dotted line segments of the image are replaced with new pixel values, where $M$ is chosen from a uniform distribution. 

\subsubsection{Parallel Lines}
\begin{figure}[ht]
\centering
\includegraphics[scale=.1512]{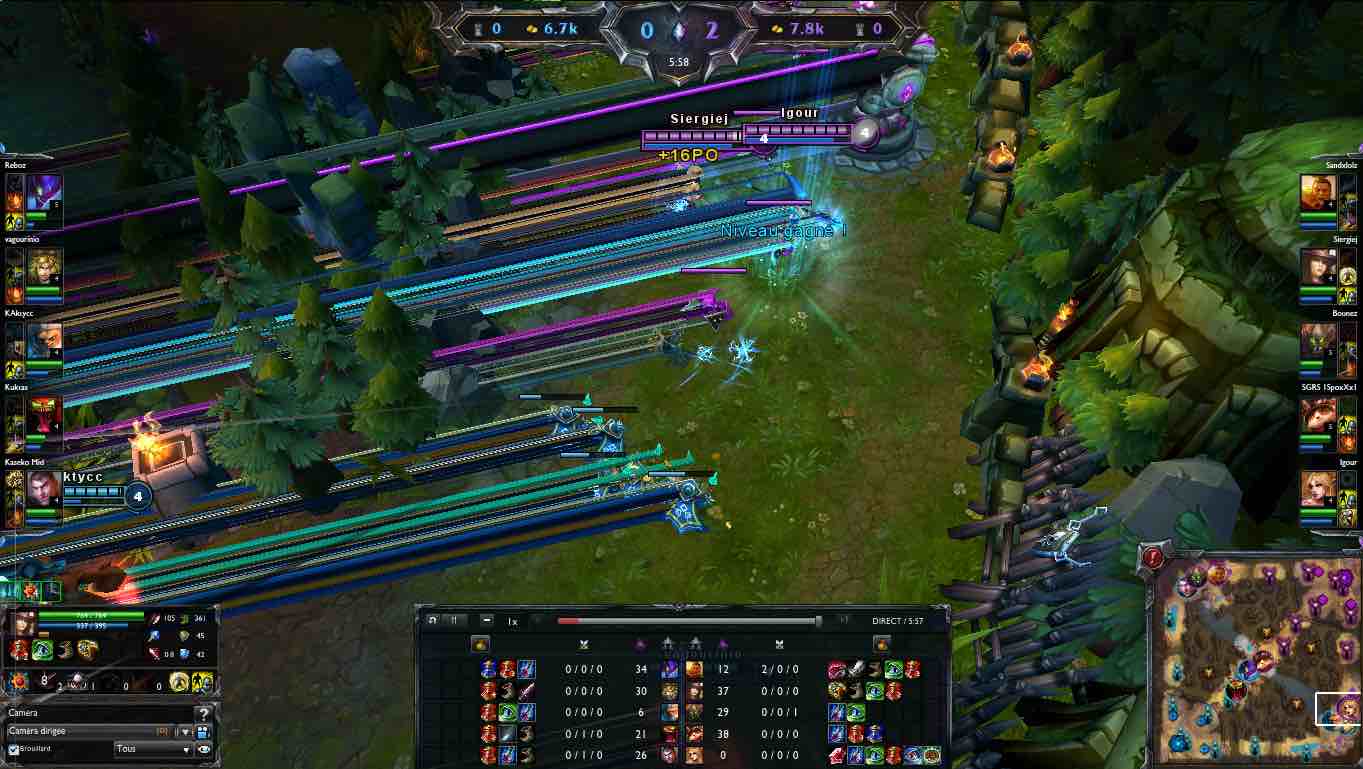}
\includegraphics[scale=.13]{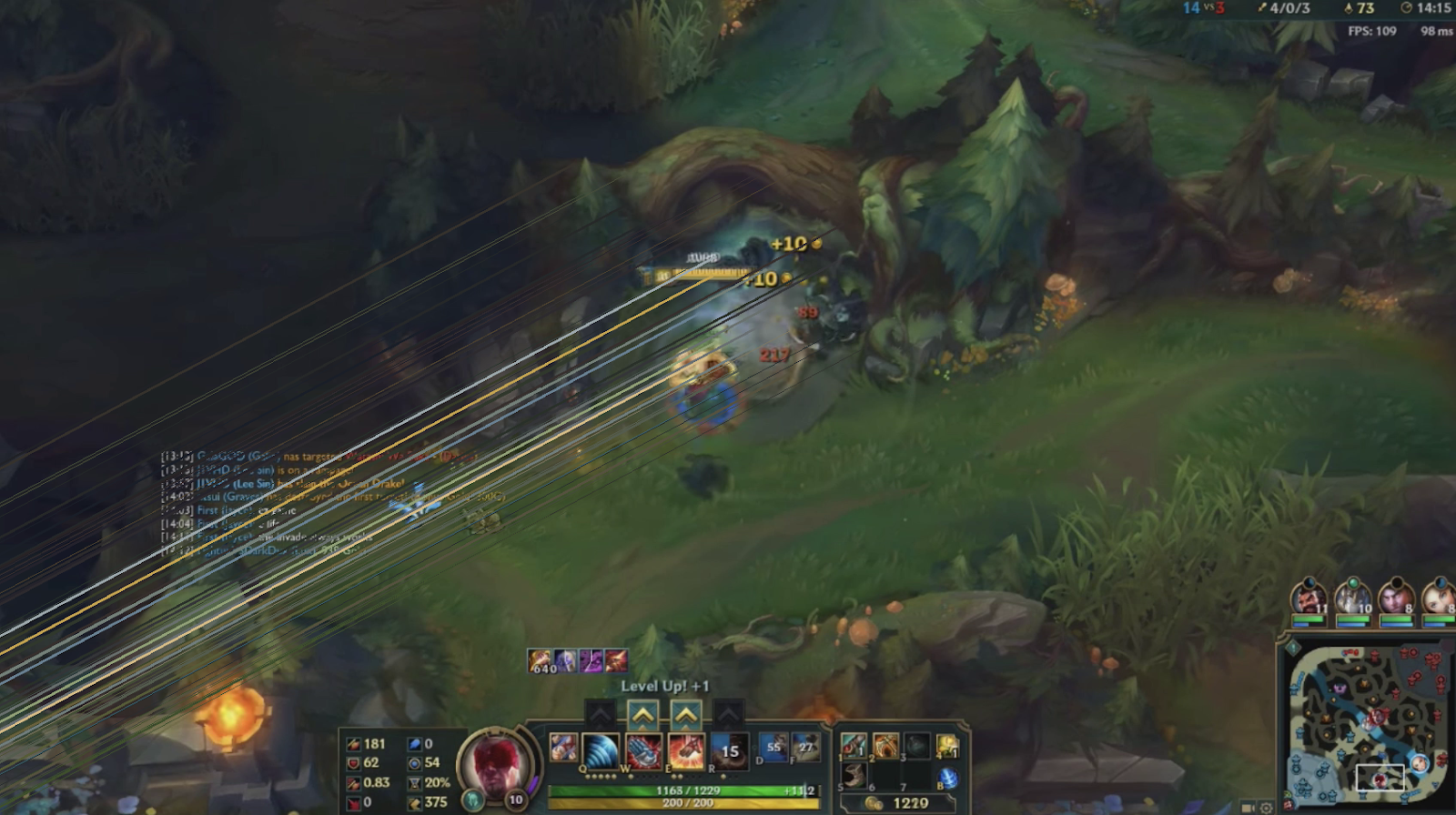}
\caption{\textit{Parallel Lines}: actual (left) and reproduced with \emph{Glitchify} (right).}
\label{fig:pl}
\end{figure}

 \textit{Parallel Lines} artifacts are visually discernible, where  the corrupted images may contain lines, where the color of each line segment is the pixel color of the starting point of the line. To reproduce the \textit{Parallel Lines} artifact (Figure \ref{fig:pl}),  $N$ parallel lines in the image are replaced by new pixel values, where $N \sim Uniform(60,100).$ The angle between each line and the horizontal axis is $\Theta$ such that $\Theta \sim Uniform(10,35)$. We choose the starting point $P = (x_0,y_0)$ of each parallel line such that $x_0 \in [0.3\cdot 1920, 0.6\cdot 1920]$ and $y_0 \in [0.2\cdot 1080, 0.8\cdot 1080]$. Each parallel line has a thickness $t$ such that $t \sim Uniform(1,3)$.

\subsubsection{Triangulation}
\begin{figure}[ht]
\centering
\includegraphics[scale=.1368]{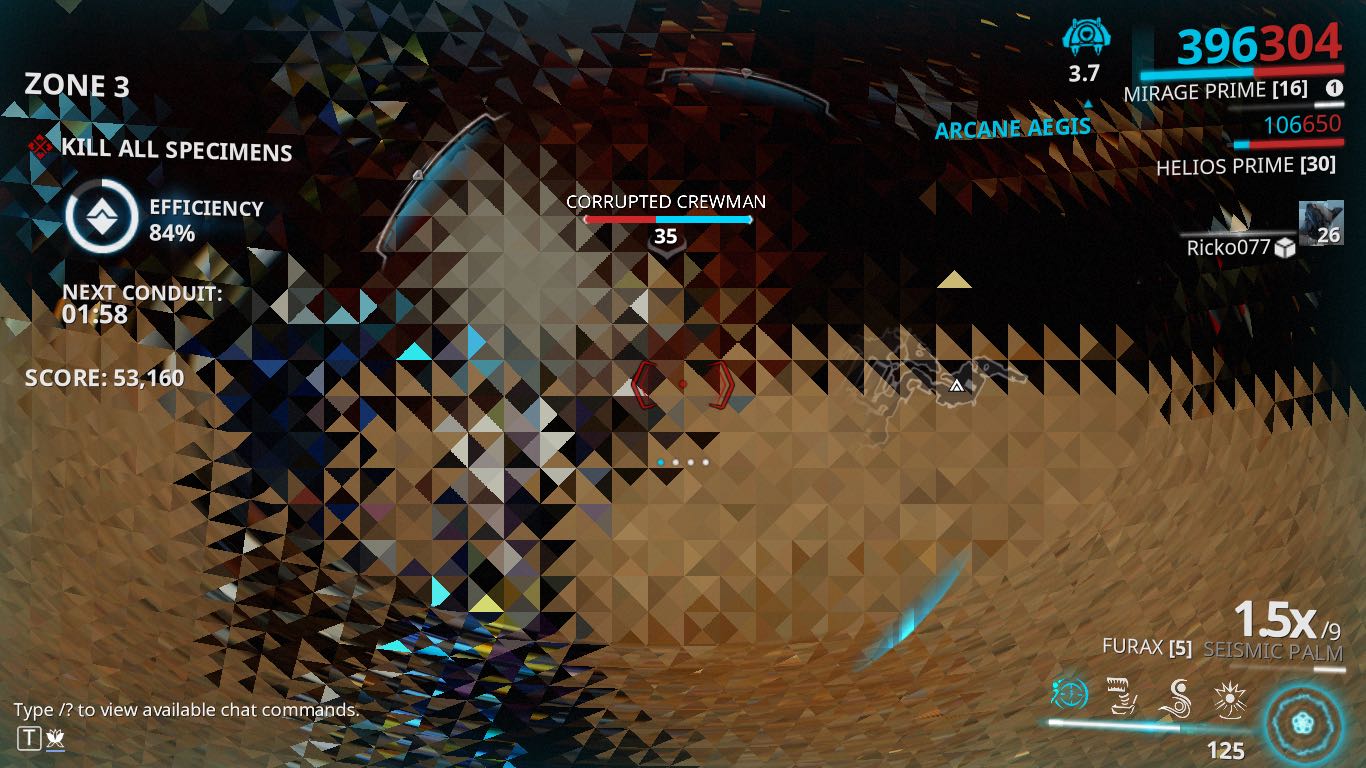}
\includegraphics[scale=.126]{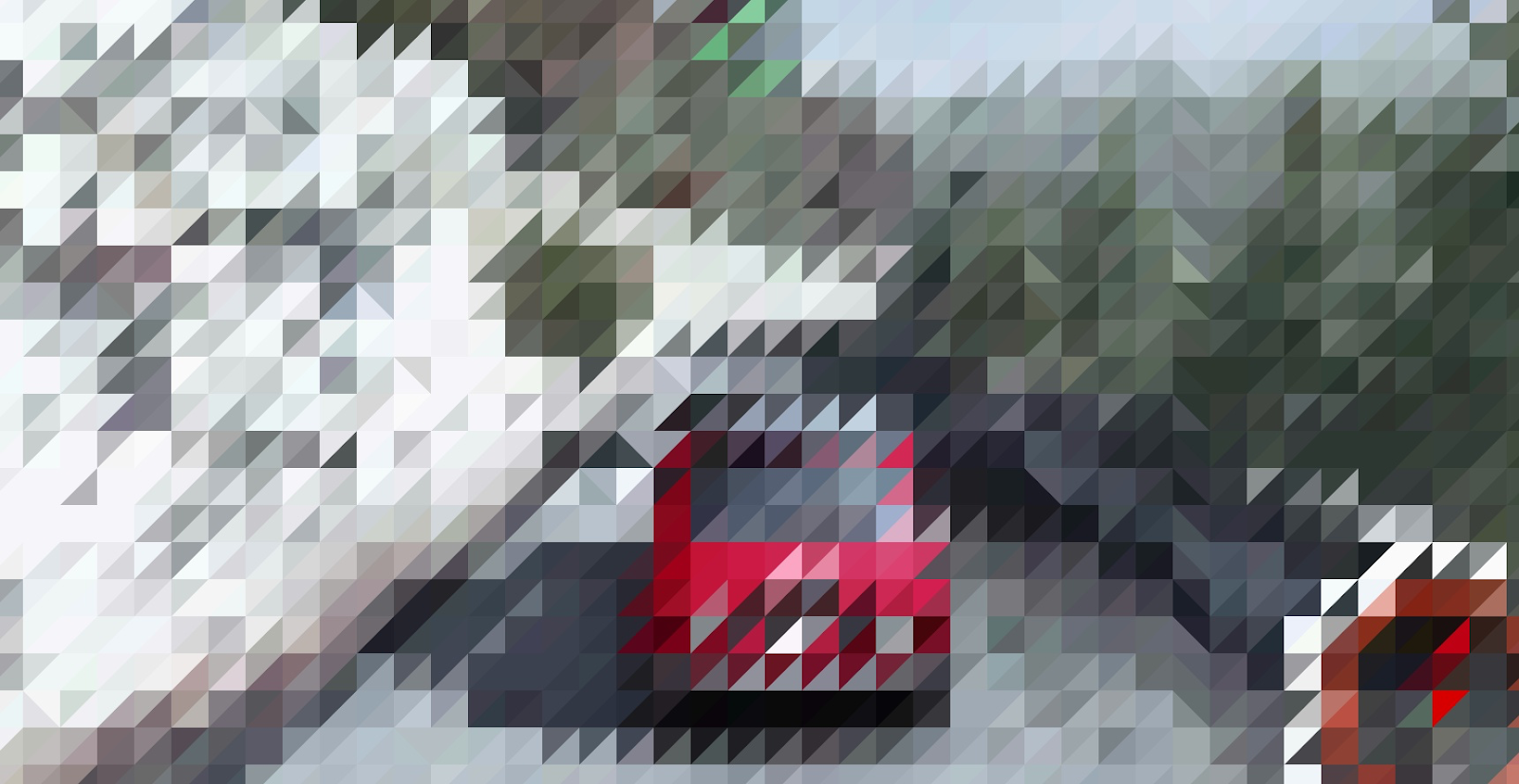}
\caption{Triangulation: actual (left) and reproduced with \emph{Glitchify} (right).}
\label{fig:tri}
\end{figure}

\textit{Triangulation} typically occurs in intensive 3D games, where surfaces are rendered by little triangles that form triangle meshes. Due to graphic defects, such triangular meshes are displayed at a coarse resolution and incorrectly colored, instead of smoothly rendered. To reproduce this artifact (Figure \ref{fig:tri}), we divide the image into triangular sections and color each triangle with the average of all the pixels within.

\subsubsection{Line pixelation}
\begin{figure}[ht]
\centering
\includegraphics[scale=0.135]{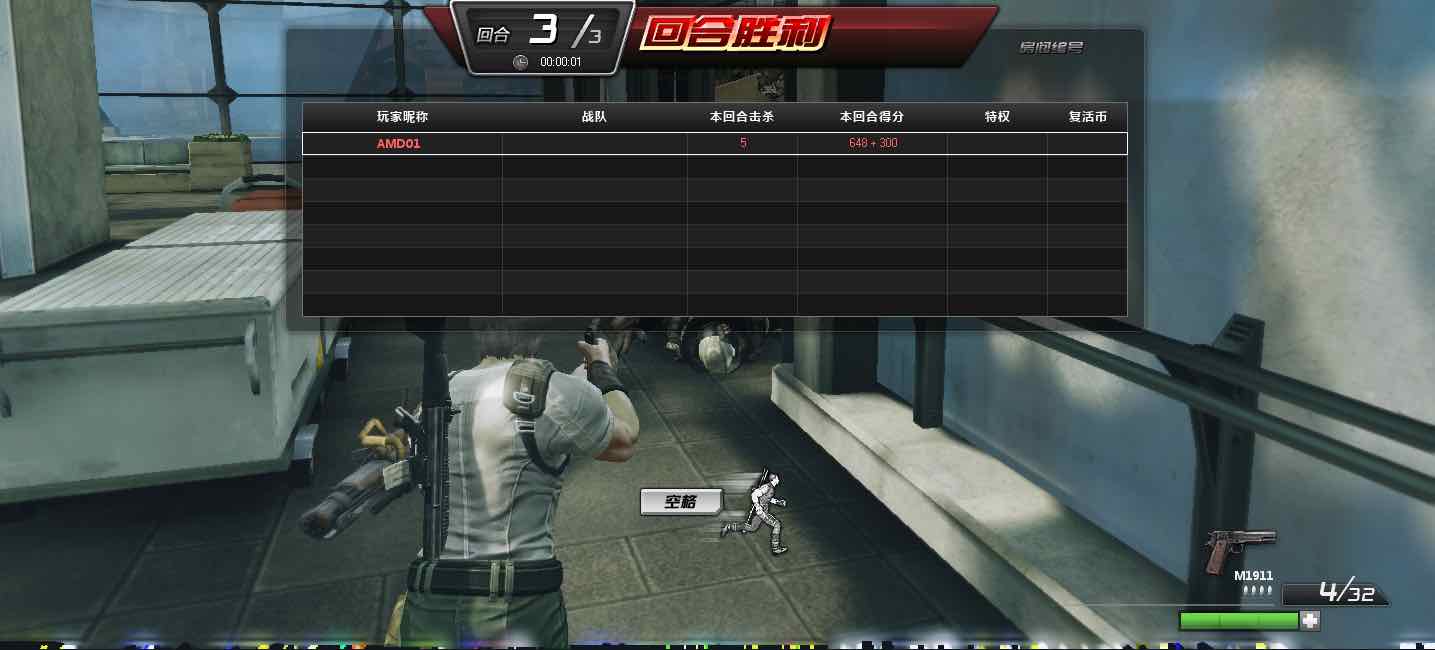}
\includegraphics[scale=.782]{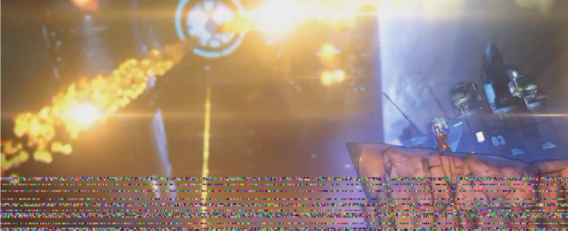}
\caption{Line pixelation: actual (left) and reproduced with \emph{Glitchify} (right).}
\label{fig:lp}
\end{figure}

\textit{Line pixelation} is characterized by noisy stripes with random orientations and positions on the image. To capture the distribution (pixel intensity variations) in this artifact, we insert a random number of noisy stripes with random orientations and positions in addition to randomly positioned halos around some pixels. This artifact appears at the very bottom of the left image in figure (Figure \ref{fig:lp}).

\subsubsection{Screen Stuttering}
\begin{figure}[ht]
\centering
\includegraphics[scale=0.216]{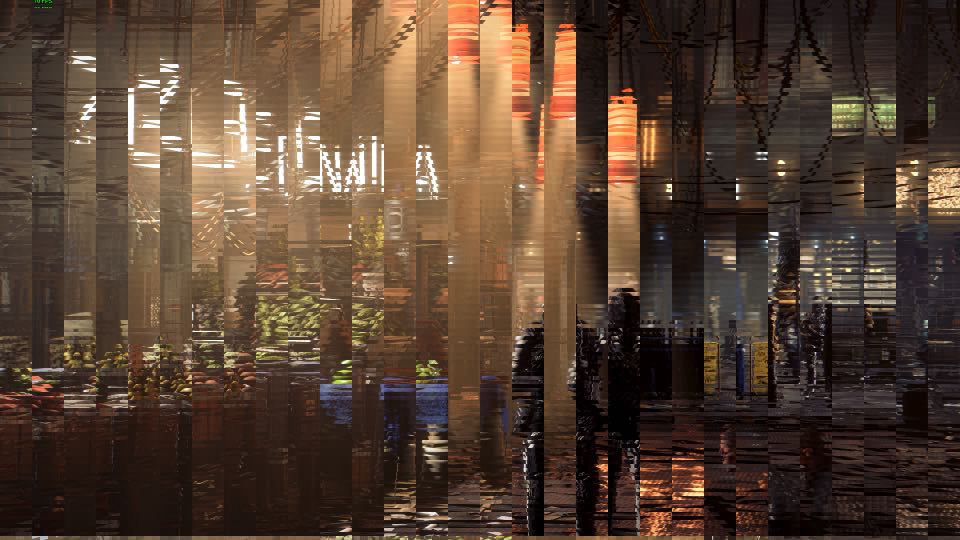}
\includegraphics[scale=0.108]{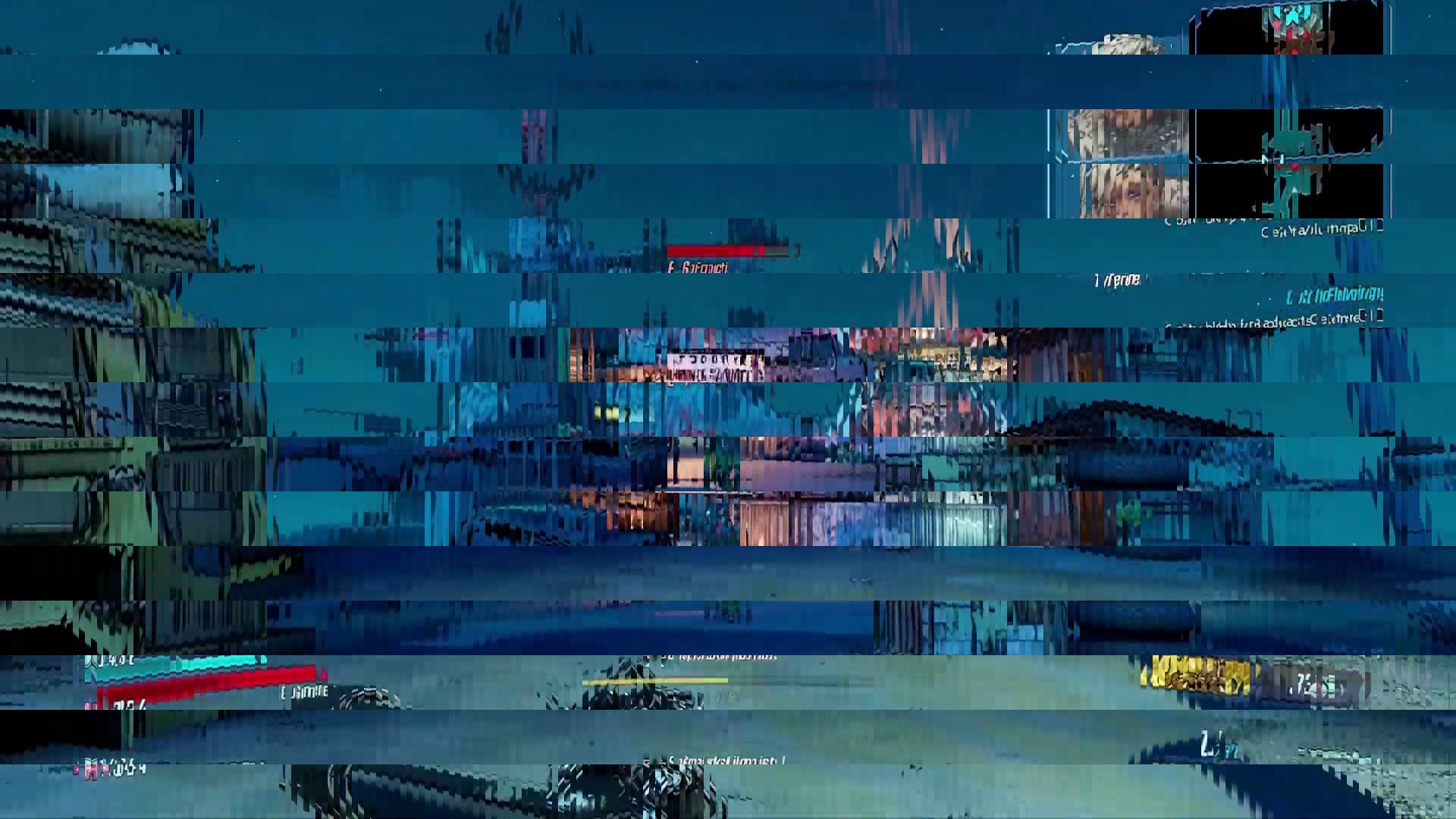}
\caption{Screen stuttering: actual (left) and reproduced with \emph{Glitchify} (right).}
\label{fig:stuttering}
\end{figure}

\textit{Screen stuttering} occurs when neighboring columns and rows of the image are swapped. We  reproduce the \textit{Screen stuttering} artifact (Figure \ref{fig:stuttering}) by swapping neighboring columns and rows in one direction, and then swapping the neighboring rows and then columns in the other direction respectively.

\subsubsection{\textit{Screen tearing}}
\begin{figure}[ht]
\centering
\includegraphics[scale=0.2844]{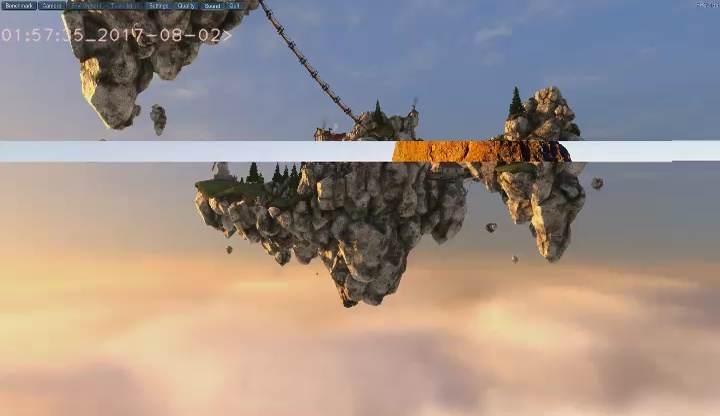}
\includegraphics[scale=0.594]{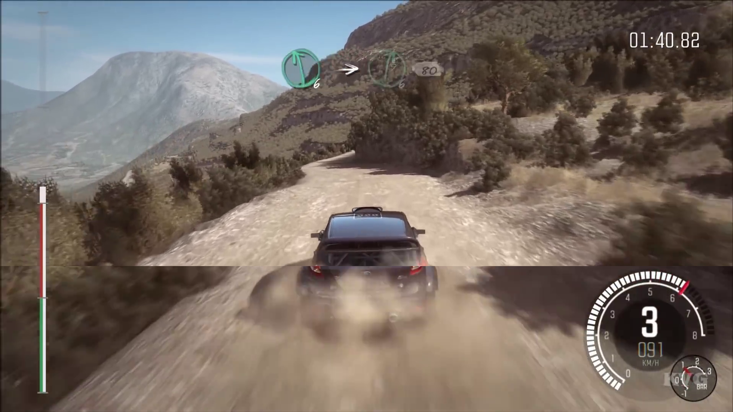}
\caption{\textit{Screen tearing}: actual (left) and reproduced with \emph{Glitchify} (right).}
\label{fig:tearing}
\end{figure}

\textit{\textit{Screen tearing}} occurs when two consecutive frames in a video are rendered in same image. Therefore, part of the image shows the scene at a certain point in time, while the other part of the same image shows that scene at a later time. To reproduce this  artifact, we select two frames in a video that are 100 frames apart from each other, and then randomly replace some rows (or columns) of the first frame with the corresponding rows (or columns) of the second frame (Figure \ref{fig:tearing}).

\section{Feature Extraction}\label{sec:feature}
Since we extracted high-quality colored images ($1920 \times  1080$ pixels), the dimensions of the images were too large to be used directly by machine learning algorithms. Therefore, the following methods were used to extract low dimensional features from images. 

\subsection{Discrete Fourier Transform}

The  two-dimensional  Discrete  Fourier  Transform  (DFT)  is  often used  to  process  two-dimensional discrete signals such as images. Given a one-channel signal $f(x,y)$ with dimensions $M\times N$, its DFT $F(u,v)$ is given by
\begin{equation}
F(u,v)=\frac{1}{\sqrt{MN}}\sum\limits_{x=0}^{M-1}\sum\limits_{y=0}^{N-1}f(x,y)e^{-2i\pi\frac{ux}{M}}e^{-2i\pi \frac{vy}{N}},
\end{equation} 
where $u$ and $v$ are the spectral coordinates.
 
\noindent The original signal can be extracted from $F(u,v)$ via inverse Fourier transform:
\begin{equation}
f(x,y)=\frac{1}{\sqrt{MN}}\sum\limits_{u=0}^{M-1}\sum\limits_{v=0}^{N-1}F(u,v)e^{2i\pi\frac{ux}{M}}e^{2i\pi \frac{vy}{N}}.
\end{equation}

\noindent The rationale behind considering DFT as a feature for the graphics artifact classification is that several types of artifacts (e.g. morse code, \textit{Parallel Lines}, etc.) exhibit periodic patterns that could be best identified in the frequency domain. Previous studies provide evidence of successful application of this technique to the detection of periodic patterns in signals \cite{russians}. Additionally, most graphics corruptions have sharp edges and fine structure, which finds reflection in the high-frequency components.

\subsection{Histogram of Oriented Gradients (HoG)}
Histogram of oriented gradients is a feature used in computer vision to detect edges in an image \cite{1467360}. An $M \times N$ color image can be represented using three functions $R,G,B : \mathbb{R}^{M \times N} \rightarrow R$ that map each coordinate to the corresponding red, green, and blue color intensity value, respectively. The gradient of the functions at each coordinate can be approximated by applying discrete derivative masks at each coordinate. \\

\noindent The image is then divided into small patches, and the magnitude and orientation of gradients within each patch are computed and summarized by a histogram of gradients containing $n$ bins corresponding to angles $0, \frac{\pi}{n}, \frac{2\pi}{n}, \ldots, \frac{(n-1)\pi}{n}$. For each gradient with magnitude $m$ and orientation $\theta$, we select the two consecutive bins (here we consider the last bin and the first bin to be consecutive) such that $\theta$ lies in the range determined by the two bins. Suppose the two selected bins correspond to angles $\theta_1$ and $\theta_2$, then the values go to the two bins are $\frac{mn|\theta-\theta_1|}{\pi}$ and $\frac{mn|\theta-\theta_2|}{\pi}$, respectively. Finally, we normalize the histograms and then concatenate them together to form a feature descriptor of the entire image.

\subsection{Pixel-wise Anomaly Measure}
\noindent Given an image, we approximate the distribution of red, green, and blue intensities, and then assign each individual pixel an anomaly score based on how much the pixel's intensity deviates from the estimated global distribution \cite{RX_detector}. 
This process can be done using graph-based method described below \cite{graph_lap} as follows.\\

\noindent Consider an undirected, weighted graph $G = (V,E)$ composed of a vertex set $V =\{r,g,b\}$ corresponding to the three color intensities, and an edge set $E$ specified by $(a, b, w_{ab})$, where $a, b \in V$, and $w_{ab} \in \mathbb{R}^{+}$ is the edge weight between vertices $a$ and $b$.  In our case, the edge weights are defined as:
\begin{equation}
w_{rg}=\frac{1}{1+\left(\frac{\mu_{r}-\mu_{g}}{\alpha}\right)^{2}}, w_{rb}=\frac{1}{1+\left(\frac{\mu_{r}-\mu_{b}}{\alpha}\right)^{2}}, w_{gb}=\frac{1}{1+\left(\frac{\mu_{g}-\mu_{b}}{\alpha}\right)^{2}}
\end{equation}
\noindent where $\alpha = \frac{\mu_a + \mu_g + \mu_b}{3}$ and $\mu_r, \mu_g, \mu_b$ are the average red, green, and blue intensities in the image, respectively. From the adjacency matrix $W$, the combinatorial graph Laplacian matrix
$L = D - W$ can be computed, where $D$ is the degree matrix defined by:
\begin{equation}
D(a, b)=\left\{\begin{array}{ll}{\sum_{k=1}^{n} w_{a k}} & {\text { if } a=b} \\ {0} & {\text { otherwise }}\end{array}\right.
\end{equation}\hspace{\fill}\\

\noindent Finally, we normalize the Laplacian matrix $
L^{*}=D^{-\frac{1}{2}}L D^{-\frac{1}{2}}
$ and define an anomaly measure for each pixel $x$ in the image by:
$$ \delta(x) = s^T L^* s,$$
 \noindent where $s \in \mathbb{R}^3$ is the color intensity of $x$.\\
 
 \subsection{Randomized Principal Component Analysis}
 Principal component analysis (PCA) is a commonly used dimensionality reduction technique in machine learning and statistics \cite{pca_reduction}. It attempts to find directions, or principal components, that maximizes the variance of projected data. Data projected onto the space determined by the first several principal components are used as a low dimensional representation of the original data matrix.\\
 
\noindent Principal components are often computed via singular value decomposition (SVD) \cite{pca_via_svd} since the principle components are exactly the normalized right singular vectors of the data matrix. However, computing the exact value of SVD takes $O(mn \min (m,n))$ time, where $(m,n)$ is the dimension of the data matrix. Therefore it is computationally infeasible to find the exact decomposition to our high dimensional data matrix. Instead, we apply randomized power iteration SVD algorithm described in \cite{randomized_svd,random_svd2}. \\
 
\section{Classification}\label{sec:classification}

\begin{figure}[ht]
  \centering
  \includegraphics[scale=0.55]{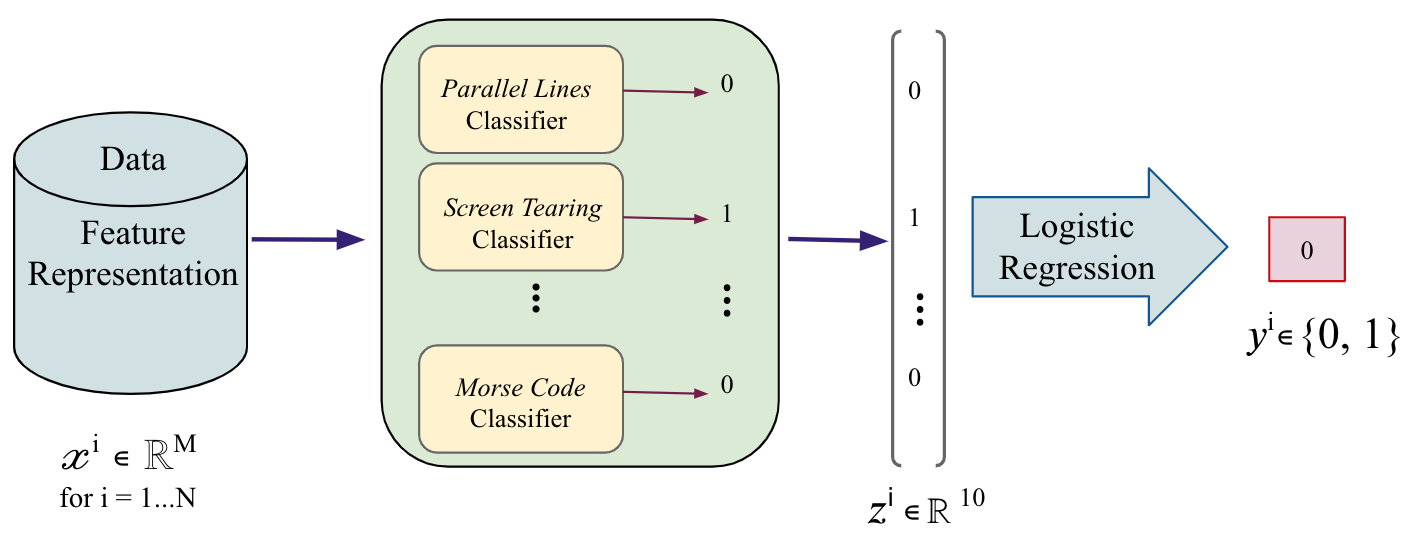}
  \caption{The Architecture of the Ensemble Model. The input consists of $N$ observations with $M$ features. The ensemble produces a binary output of ``normal" or ``corrupted''.}
\label{fig:ensemble}
\end{figure}

For each artifact, we explored various combinations of feature representations and classification algorithms, and picked the best performing combination as a specialized classifier for that artifact. We then combined these specialized classifiers into an ensemble model as they are shown to perform better than any single individual classifier (figure \ref{fig:ensemble}) \cite{why_ensemble}. The following section provides a brief overview of the justification behind the feature/model combinations that we hand selected.

\subsection{Feature and Model Selection}
Given the nature of each artifact, we only considered those feature representation/classifier combinations that seemed capable of capturing that artifact. For example, there are no repetitive patterns in \textit{Screen tearing} artifacts so we did not use Fourier transform to represent this artifact. The set of potential classifiers consisted of Convolutional Neural Networks (CNN), Logistic Regression (LR), Random Forest (RF), Support Vector Classifier (SVC), and Linear Discriminant Analysis (LDA). We used accuracy, recall, and precision in that order as evaluation metrics of our models. Additionally, if two models had similar performances, we picked the one with the lowest training time. It is important to note that if in the process of testing a feature/classifier combination on a given artifact we reached accuracy, recall, and precision rates higher than \%90 on the test set, we did not try other combinations on that artifact.

\begin{itemize}
    \item \textbf{Resize} $\rightarrow$ \textbf{CNN}: Convolutional Neural Networks are deep learning classifiers for computer vision tasks. CNNs are typically not capable of performing well on large images, so we had to resize the images. Our CNN had the architecture of Convolution $\rightarrow$ Maxpool $\rightarrow$ Convolution $\rightarrow$ Maxpool $\rightarrow$ Softmax. We did not perform hyperparameter tuning, however, it is a promising future direction to pursue. We tried this combination on artifacts that were still detectable after resizing: \textit{Parallel Lines}, \textit{Shapes}, \textit{Shader}, and \textit{Discoloration}. For the artifacts with a repetitive pattern, we added a Fourier transform step prior to resizing the image as the transformed image will have distinctive features that are more dominant than in the original image, especially after the image has been resized. The \textbf{Fourier Transform} $\rightarrow$ \textbf{Resize} $\rightarrow$ \textbf{CNN} combination was used on every artifact except for \textit{Parallel Lines}, \textit{Screen Tearing}, and \textit{Discoloration}.
    \item \textbf{Fourier Transform} $\rightarrow$ \textbf{Resize} $\rightarrow$ (\textbf{PCA}) $\rightarrow$ \textbf{SVC, LDA, LR}: Given that the classification models in this category are shallow and do not take in images as input, we either used PCA as a dimensionality reduction technique, or we flattened the image into a one dimensional vector before handing the to the classifiers. Also, since our transformed images were still too large, we downsized the images before the application of PCA or flattening. Fourier transform was used as a first step because in most artifacts, the transformed image has more distinctive and less localized visual features than the original image. We used this combination on every artifact except for \textit{Screen Tearing} and \textit{Discoloration}.
    \item \textbf{Resize} $\rightarrow$ (\textbf{HOG}) $\rightarrow$ \textbf{SVC, LDA, LR}: This is similar to the above combination, with the difference that HOG is used instead of PCA. We tried this combination on artifacts that would have distinctive straight edges in their appearance which HOG is good at capturing: \textit{Screen Tearing}, \textit{Shapes}, \textit{Shader}, \textit{Line Pixelation}.
    \item \textbf{Anomaly Measure} $\rightarrow$ (\textbf{HOG}) $\rightarrow$ \textbf{Threshold}: This computationally cheap combination produced good results on artifacts that typically have sharp color contrasts with the neighboring pixels. We tried this on all artifacts except for \textit{Triangulation}, \textit{Screen Tearing}, and \textit{Discoloration}.
\end{itemize}

 Table \ref{tab:models} illustrates the best performing feature representation and classifier combinations that were used for each artifact. We discovered that even though CNNs and random forests are widely and successfully used in computer vision and classification tasks respectively, Logistic Regression outperforms them in most of our artifact detection tasks, especially given their relatively low training time. We used Logistic Regression model provided by the sklearn package\cite{scikit-learn} and used their default arguments; that is $L2$ norm for penalization and a regularization strength of $c=1.0$.  As with LR, we used the sklearn SVC model (RBF kernel) to perform binary classification of the extracted features.

 \section{Training and Testing of the Ensemble}\label{sec:ensemble}
 Here we describe the data used for our experiments and the training process of the ensemble in three stages. In training stage I we trained the specialized classifiers for each glitch, that is, the components of our ensemble just like any other ensemble. For combining the results of the specialized classifiers, using the OR logic - which is the most basic combining method - did not seem reasonable for our application, therefore we used a Logistic Regression instead \cite{LR_ensemble_1, LR_ensemble_2}. That is because, if some of our specialized classifiers happened to have a pattern in false classifications - for example when the \textit{Screen tearing} and \textit{Shader} models tend to label a normal image as corrupted while no other specialized classifier would - the Logistic Regression has a much better tendency to capture that than the OR logic. Therefore, we used training stage II to train the ensemble Logistic Regression. Finally, to test the generalizabilty of our classifier to new games, we tested the ensemble on a dataset consisting of games that have never been seen by either the specialized classifier or the ensemble Logistic Regression. These three stages are explained in detail in the following paragraphs.
 
 \subsection{Data}
 
We applied \emph{Glitchify} to generate the artifacts described previously, to obtain  a large dataset for training. The initial, uncorrupted images were collected by downloading long (2-6 hour) gameplay videos from 30 different games. We then extracted around 2,000 images from each game until we had a total of 50,000 images. After extracting all the  images, we visually inspected all of the images to make sure they appeared glitch-free. We then added all the 12 different types of artifacts to half of the images, leaving the other half unchanged and labeled them as normal (glitch-free) images. 

We  split this dataset of 50,000 images into three: A, B and C consisting of approximately 35,000, 7,500 and 7,500 images respectively . Dataset A consisted of gaming images from 24 games and was  the largest dataset of the three. It was used to train specialized classifiers. Each specialized classifier was responsible for capturing only one type of graphic corruption. Dataset B contains gaming images from 3 games distinct from those in Dataset A, and it is used to train a Logistic Regression model that combines the outputs from the specialized classifiers and makes a final prediction on whether the input image is corrupted or not. Dataset C is the holdout dataset which is reserved for testing of the ensemble model. All of the data

\begin{table}[ht]
\centering
\begin{tabular}{@{}llll@{}}
\toprule
  & \multicolumn{1}{c|}{Game} & \multicolumn{1}{c|}{Dataset} & \multicolumn{1}{c}{Stage of Usage}\\
 
 \midrule
1 & Ancestor Legacy &  &  \\
2 & Control &  &  \\
3 & Detroit: Become Human &   &  \\
4 & Devil May Cry 5 &  & \\
5 & Dirt Rally 2 &  & \\
6 & Far Cry 5 &  & \\
7 & Final Fantasy  & & \\
8 & Hollow Knight (2d) & & \\
9 & GTA & & \\
10 & Overcooked & & \\
11 & League of Legends  & & \\
12 & Assassin's Creed & & \\
13 & Star Control: Origins & A  & Training Stage I \\
14 & Total War: 3 Kingdoms & & \\
15 & Star Craft 2 & & \\
16 & Unravel 2 & & \\
17 & Kingdom Come & & \\
18 & Metro Exodus  & & \\
19 & Need For Speed Payback & & \\
20 & Mutant Year Zero & & \\
21 & Rage 2 & & \\
22 & Resident Evil & & \\
23 & Strange Brigade  & & \\
24 & The Sinking City & & \\
\midrule
25 & Battlefield  & & \\
26 & DOTA 2 & B & Training Stage II\\
27 & Hob & & \\
\midrule
28 & Civilization 6 & & \\
29 & Crackdown 3 & C & Ensemble Testing\\
30 & Forza Horizon  & & \\
\bottomrule
\end{tabular}
\caption{Best performing feature representation/classifier combination for each artifact.}
\label{tab:games}
\end{table}

\subsection{Training Stage I}
During this stage, we trained the specialized classifiers. For this training procedure, we extracted images from 24 games (dataset $A$ in Table \ref{tab:games}), labeled 2400 of these images as normal (artifact-free) and applied \emph{Glitchify} to the rest, obtaining around 1,500 corrupted frames per artifact. It should be noted that the training sets for each of the specialized classifiers were mutually exclusive (they did not share any normal images in common). A variety of models introduced in sections \ref{sec:classification} were paired with features from section \ref{sec:feature} and trained as binary classifiers for each artifact type. We used a $50/50$ train-test split of the data. The best-performing feature/model combinations were chosen based on the accuracy and recall and are recorded in table \ref{tab:models}. The performance of these specialized classifiers on the test set (i.e. on familiar games) is shown in figure \ref{fig:familiar}.

\subsection{Training Stage II}
After training the specialized classifiers, we form an ensemble by  training an LR model that takes the concatenated outputs of the specialized classifiers as input, and outputs a 0 (normal) or 1 (corrupted). To ensure generalizability across games, our dataset for this training stage comprised of 3 games (dataset $B$ in Table \ref{tab:games}) that were \emph{never seen} by the specialized classifiers. The dataset for this stage consisted of 1650 normal images and 150 images of each type of artifact. A $75/25$ train-test split was applied to the data. At this stage, all of the specialized classifiers are shown normal images in addition to all types of artifacts. The performance of the specialized classifiers on the test set (i.e. on new games) is reported in figure \ref{fig:new}. Given that the real-world application of this ensemble is expected to work well on games that have not been seen before, we did not train the Logistic Regression on familiar games.

\subsection{Testing the Ensemble}
The goal of this stage is to measure the generalizability of the ensemble model as a whole across different games. The dataset for this testing stage comprised of 3 games that are never seen by the ensemble (dataset C in Table \ref{tab:games}), with 1650 normal images and 1800 corrupted images, which is 150 for each artifact type. The results are reported in figure \ref{fig:LR}.

\begin{table}[ht]
\centering
\begin{tabular}{@{}lll@{}}
\toprule
Artifact & Feature Representation & Classifier  \\ \midrule
\textit{Shapes} & FT + Resize & LR \\
Line pixelation & Anomaly Measure + Dilation & Threshold \\
\textit{Shader} & Resize + HOG & LR\\
Morse code & FT + Resize & SVC\\
\textit{Parallel Lines} & FT + Resize + PCA & LR \\
Dotted line & FT + Resize + PCA & LR\\
Stuttering & FT + Resize + PCA & LR\\
Triangulation & FT + Resize + PCA & LDA \\
\textit{Discoloration} & FT + Resize + PCA & LR \\
\textit{Screen tearing} & Resize + HOG & LR \\

\bottomrule
\end{tabular}
\caption{Best performing feature representation/ classifier combination for each artifact}
\label{tab:models}
\end{table}

\section{Results}\label{sec:results}

\subsection{Evaluation Metrics}
To evaluate the performance of our models, we used three different metrics: accuracy, precision and recall. We refer to corrupted images as \emph{positive} and normal images as \emph{negative}. 

\subsection{Performance of our Model on Familiar Games}
\begin{figure}[ht]
  \centering
  \includegraphics[scale=0.6]{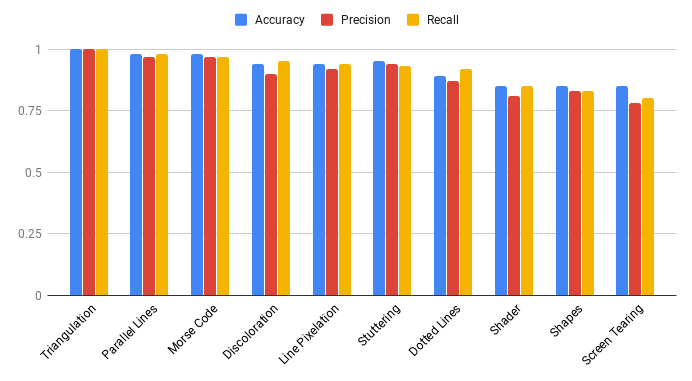}
\caption{The performance of individual classifiers on a test set consisting of familiar games. }
\label{fig:familiar}
\end{figure}

\begin{figure}[!h]
  \centering
  \includegraphics[scale=0.5]{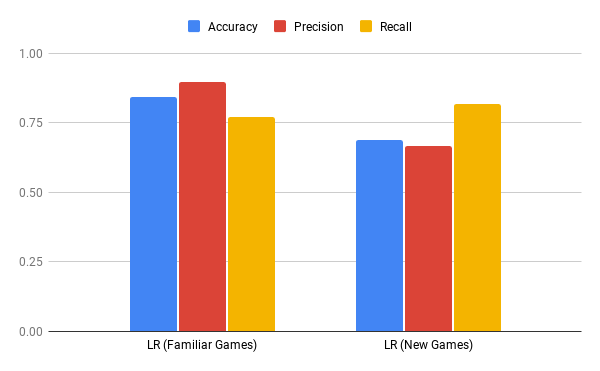}
\caption{Ensemble logistic regression on new vs. familiar Games}
\label{fig:LR}
\end{figure}

We first tested the performance of the specialized classifiers and the ensemble model on familiar games. Figure \ref{fig:familiar} illustrates the performance of the specialized classifiers on familiar games where we used the held out test set from training stage I (dataset A). Figure \ref{fig:LR} illustrates the performance of the ensemble Logistic Regression on familiar games where we used the heldout test set from training stage II (dataset B). 

\subsection{Performance of our Model on New Games}

An important metric in assessing artifact detection models is generalizability. In our case, generalizability refers to the ability of the model to perform well on images extracted from games that have not been encountered before. We will measure both the generalizability of the ensemble Logistic Regression and that of the specialized classifiers.

\begin{figure}[!h]
  \centering
  \includegraphics[scale=0.45]{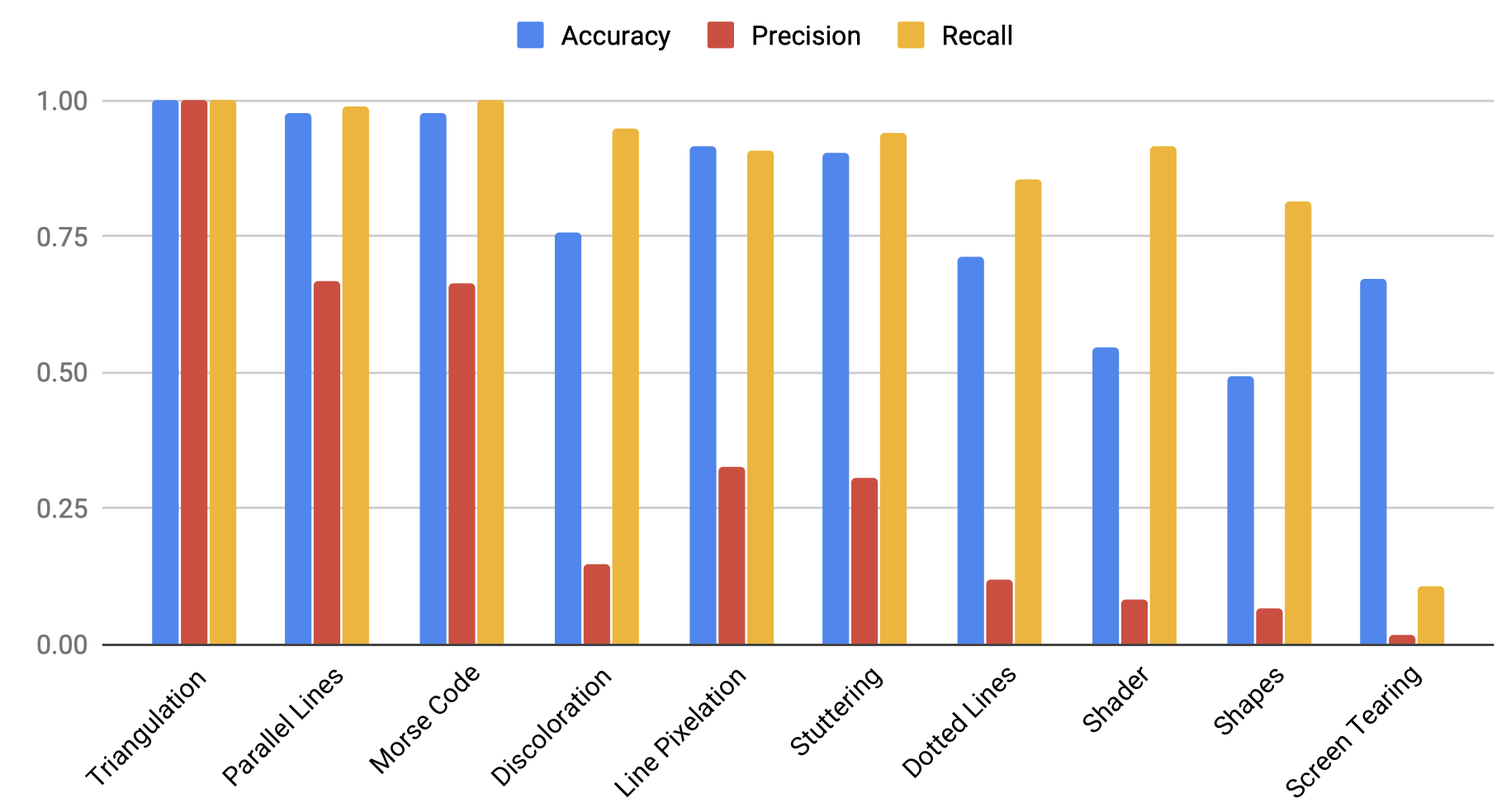}
\caption{The performance of individual classifiers on a test set consisting of new games. }
\label{fig:new}
\end{figure}

In training stage 2, even though the Logistic Regression model is trained using the games in dataset B, the specialized classifiers have never seen any image from dataset B. Therefore we used images from dataset B to test and evaluate the generalizability of the specialized classifiers. We also tested the generalizability of the ensemble model on dataset C which neither the ensemble model nor the individual classifiers have seen before, and obtained an accuracy of 69\%. Figure \ref{fig:new} shows the testing results of each individual classifier.

\section{Discussion}

\begin{figure}[ht]
\centering
\includegraphics[scale=0.5]{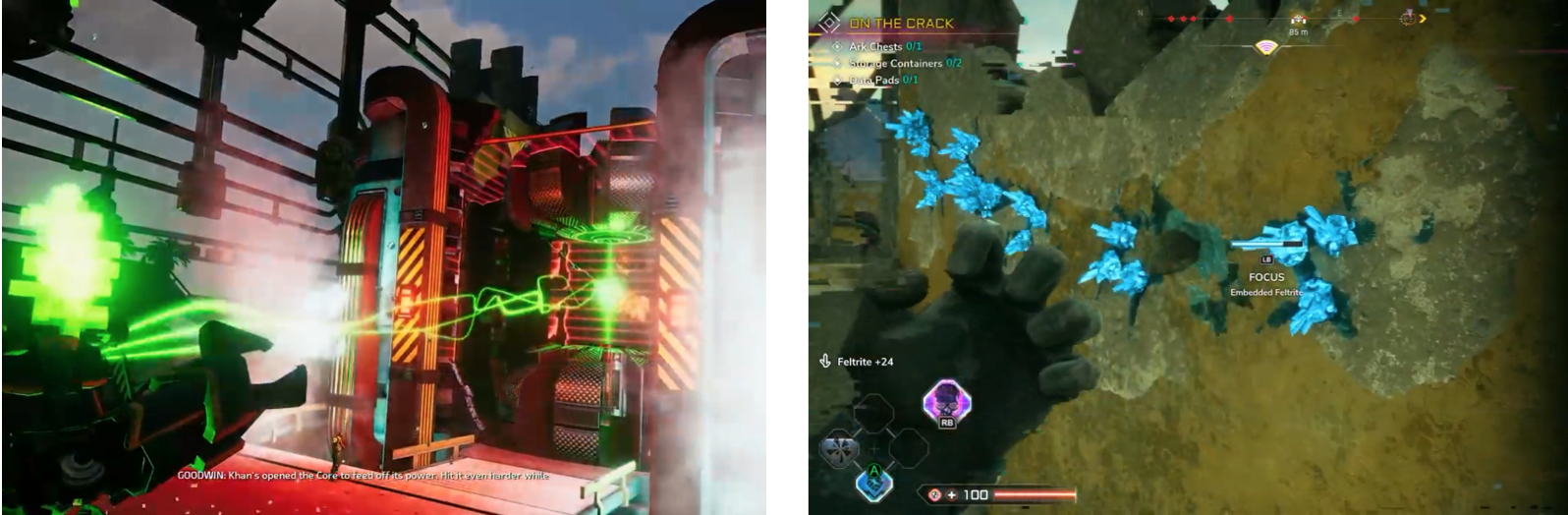}
\caption{Normal images that look corrupted when taken outside the context of the game. The image on the left has stuttering patterns in the top left section of the frame. The image on the right has green glitchy patterns in the left section of the frame.}
\label{confused}
\end{figure}

We now discuss possible interpretations of our results in addition to their limitations and sources of possible bias. One fundamental concern relates to our generated dataset and its representation of the actual graphics corruptions that occur during gameplay. First, our approach in developing the \emph{Glitchify} program consisted of classifying artifacts into categories based on their appearance, however, the viability of this procedure is debatable.
Although the structure and design of \textit{Screen tearing} and stuttering were sufficiently understood and formally expressed, our definition of \textit{Shader}/\textit{Shapes} and \textit{Line pixelation} artifacts, however, might be too narrow to capture all of the variations of these corruptions seen in reality. This discrepancy might be responsible for some bias present in our artificial dataset, which in turn affects all further results.
Second, another data-related concern comes from the frames we extracted from games to feed into \emph{Glitchify}. We require that all these frames are normal, i.e. do no contain any corruptions, before the application of \emph{Glitchify}. To ensure this holds, we manually processed collected images in an attempt to perform a first-order quality check. Due to surrealistic contents of most games, working with individual frames rather that a continuous and dynamic gameplay makes it harder, if not impossible, to correctly discriminate between unwanted artifacts and intentional design. Figure \ref{confused} displays examples of such images. 

\noindent

\begin{figure}[ht]
    \centering
    \includegraphics[scale=0.6]{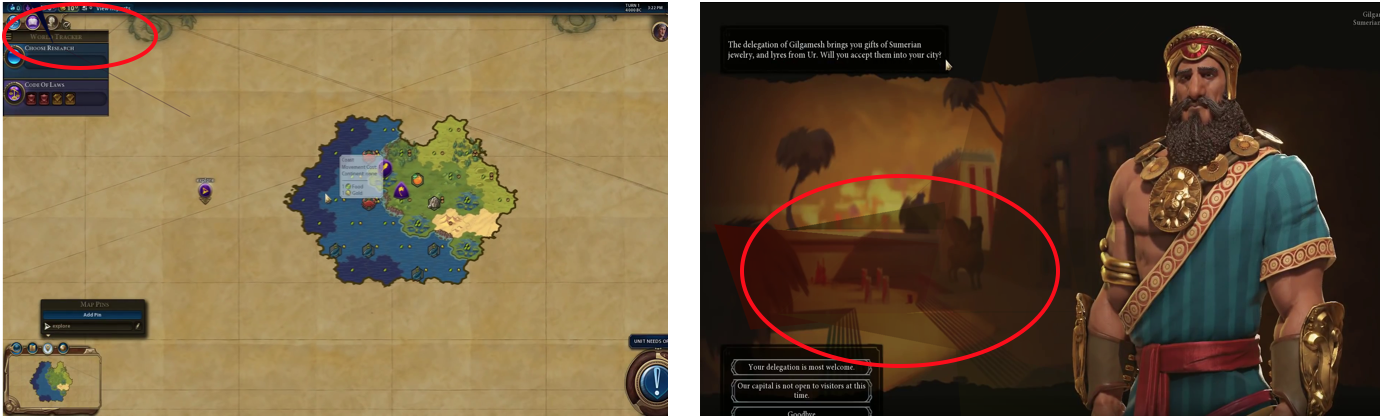}
    \caption[Visualization of False Negatives]{Visualization of False Negatives. Both of these images are corrupted through \emph{Glitchify}. The left image has a \textit{Shapes} artifact which is subtly visible in the top left corner inside the red circle. The right image has a \textit{Shapes} artifact which is hardly visible inside the red circle.}
    \label{fig:FN}
\end{figure}

Regarding the individual models, the accuracy on the test set drops from training stage 1 (figure \ref{fig:familiar}) to training stage 2 (Table \ref{fig:new}). This is mainly due to the fact the models in training stage 2 have never seen the games they are being tested on. In other words, the number and variation of the games in dataset A were not sufficient to ensure generalizabilty of the specialized classifiers across different games. One season for getting a high number of false negatives (contributing to low precision and accuracy) is that the artifacts that were added through \emph{Glitchify} are too subtle, as we can see in figure \ref{fig:FN}.\\

\begin{figure}[ht]
    \centering
    \includegraphics[scale=0.5]{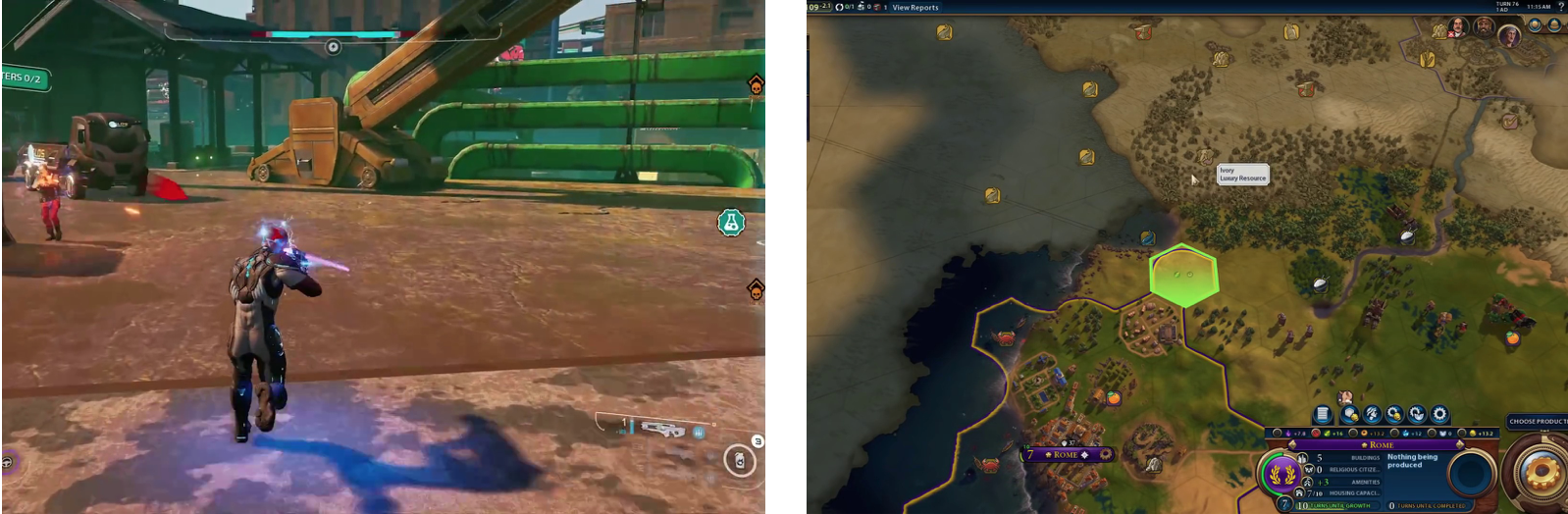}
    \caption[Visualization of False Positives]{Visualization of False Positives. On the left, we have a horizontal separation line, causing the \textit{Screen tearing} classifier to label this image as ``corrupted". On the right, we have a green color gradient (in form of a hexagonal), causing the \textit{Shader} classifier label this image as ``corrupted".}
    \label{fig:FP}
\end{figure}

In this application, however, the recall rate is the most important metric to consider as it is crucial to make sure that during gameplay, we detect as many corrupted images as possible so we can fix them. In training stage 2, we can see that most models produce good recalls on games they have never seen, showing a good degree of generalizability. The artifacts comprising \textit{Discoloration} and \textit{Screen tearing} have relatively low recall scores in training stage 2 (0.66 and 0.5 respectively) versus 0.95 and 0.80 in training stage 1, demonstrating that those models overfit the training data and are not generalizable to new games. These two artifacts are especially challenging, because as we can see in figure \ref{fig:FP}, it is possible for images to have a natural separation line or color gradients which will end up confusing the \textit{Screen tearing} and \textit{Discoloration} classifiers respectively.

We speculate that the low accuracy score we have obtained on the heldout test set is due to the ``glitchy" look of one of the games we have included in dataset C (Crackdown 3). 

\section{Conclusion}
In this proof of concept study, we developed a set of algorithms and software that automatically detects graphics corruption in frames from video games. Based on a sample of screen corruption examples provided by AMD, 10 of the most common content-unrelated artifacts were selected, described, and then recreated with the \emph{Glitchify} program. With the help of \emph{Glitchify}, a dataset of 50,000 images was created by adding 10 different types of artifacts to normal frames extracted from 34 modern video games. Each of the 10 forms of corruption was used to train a basket of models that included Logistic Regression, Support Vector Machines, and Linear Discriminant Analysis. The output probabilities of these individual classifiers were used for training a mixed experts Logistic Regression model intended to carry out the final classification decision. The overall accuracy on the \emph{Glitchify}-produced test set based on games unseen during training is 69\%. 

Overall, this study has demonstrated several important results. The accuracy of the models trained on \emph{Glitchify} indicates that synthetic generation of defects can be an efficient mechanism to train models to identify real corruption. Put another way, the simple models for synthetic corruption described previously accurately represented some of the actual defects that occur in software or hardware. This hints that synthetic data can be used to efficiently train machine learning models for visual corruption. This result should be explored more generally. If true, this is an important result, because this is an effective means of generating labelled data \emph{en masse}, and circumventing the need for large, labelled datasets. 

Another important contribution of this work is demonstrating that a basket of models can be used as an efficient classifier of visual defects. No single model outperformed the others for identifying all forms of corruption. The efficacy of 'mixture of experts' approaches are well-understood in other domains in machine learning, but to the authors' knowledge, have not been previously demonstrated in the visual corruption domain.
We note that Generative Adversarial Networks could be used in the future for artifact creation. We did not use them in this work  owing to the  limited and small number of real world artifact examples available to us, as GANs are computationally expensive and require large amounts of data to achieve superior performance.

There are a few points that were not the main focus of this paper but could improve the results. First, we found that contrary to our assumptions, LR (a shallow model) outperformed CNN (a deep learning model for image classification) in most if not all of the cases. We expect that a larger dataset for the purpose of training in addition to a more exhaustive hyperparameter tuning would prove benefical to the use of CNNs. Second, we did not perform any significance testing to see which feature/model combination performs the best. Adding this step to analysis would certainly make it more rigorous. Lastly, we did not consider the possibility of having multiple kinds of artifacts in the same image. For example, it is possible to have \textit{Stuttering} and \textit{Screen Tearing} appear in a single frame, but it was not among the primary goals of this paper to be able to catch such artifact combinations.  While we realize this is a preliminary work, we believe that this work is a good starting point that can help facilitate future research on the topic of automating artifact detection which would in turn lead to significant quality improvement of video games followed by an increase in gaming revenue. In addition to individual frames, the model we have created is able to work on videos and capture glitches in real time gameplay as we demonstrate in a short demo here: \url{https://youtu.be/AiZ0Dae7jW4}. Further work needs to be done in order to improve the accuracy of our classifier as well as potentially fixing the artifact before it is displayed to the user.

\bibliographystyle{unsrt}  
\bibliography{references}

\end{document}